# An approach to visualize the course of solving of a research task in humans

Vladimir L. Gavrikov[a],*, Rem G. Khlebopros[b]

[a]V.P.Astafiev Krasnoyarsk Pedagogical University, 89 A. Lebedeva St., 660049, Krasnoyarsk, Russia

[b]Research Center of Extreme States of Organism at Krasnoyarsk Scientific Center of RAS, Akademgorodok 50, 660036, Russia

Suggested running head: COMPUTER-BASED RESEARCH TASK

*Corresponding author. Post OB 8745, Akademgorodok, Krasnoyarsk, 660036, Russia, tel. +7 391 249 8402, mobile +7 913 042 4304, fax +7 391 211 07 29, E-mail address: gavrikov@kspu.ru

**Abstract**

A technique to study the dynamics of solving of a research task is suggested. The research task was based on specially developed software Right-Wrong Responder (RWR), with the participants having to reveal the response logic of the program. The participants interacted with the program in the form of a semi-binary dialogue, which implies the feedback responses of only two kinds - "right" or "wrong".

The technique has been applied to a small pilot group of volunteer participants. Some of them have successfully solved the task (solvers) and some have not (non-solvers). In the beginning of the work, the solvers did more wrong moves than non-solvers, and they did less wrong moves closer to the finish of the work. A phase portrait of the work both in solvers and non-solvers showed definite cycles that may correspond to sequences of partially true hypotheses that may be formulated by the participants during the solving of the task.

**Introduction. Definitions**

The goal of our study was to shape an experimental approach that allowed one to record the human actions while solving a research task. Also, we were interested in finding parameters of the dynamics that are characteristic of the very process of the task solving. Therefore, the main focus of the study was on the "anatomy" of the solving process, which is somewhat different from testological approaches of applied psychology that are mainly interested in a final estimation of human performance.

Our approach lies within the frame of cognitive science that has been developed since the late 50-ties. As early as in 1958, H. Simon and A.Newell (Simon, Newell, 1971; Newell, Simon, 1972) sketched a program to study problem solving in humans and put their primary interest "on process – on how particular human behaviors come about..." (Simon, Newell, 1971, p. 146). In that program, they, among other points, enumerated a search "for new tasks (e.g. perceptual and language tasks) that might provide additional arenas for testing the theories and drawing out their implications." (idem, p. 146).

Since that time, a lot of studies have been done the core notion of which was problem solving. To mention a tiny portion of the studies, correlations of problem solving and other individual characteristics have been widely studied, for instance, intelligence (Leutner, 2002), spatial thinking ability (Adeyemo, 1994), extroversion-introversion and masculinity-femininity (Kumar, Kapila, 1987), cognitive styles and reasoning ability (Antonietti, Gioletta, 1995). Beginning with 70-80ies of the 20$^{th}$ century, a use of computer-based assessment of problem solving has been a regular method to study problem solving (e.g. Baker, Mayer, 1999; Baker, O'Neil, 2002) but it is especially interesting that observation protocols can be compared with clickstream data of the same participant (Chung, de Vries, Cheak, and Stevens, Bewley 2002). Sweller (1988) developed a critical viewpoint discussing effects of problem solving on learning.

In the following, a few terms that are important for our study are outlined.

*Research task.* Deaner et al. (2006) have performed one of the most complete literature revisions on intelligence study in non-human primates. In the review, the authors differentiate between nine paradigms and within the paradigms up to thirty various procedures. It is also widely accepted that the types of problems may be i) well defined vs. ill defined and ii) novel vs. familiar. A combination of these two oppositions provides four types of possible problems: well defined and novel, ill defined and familiar and so on.

We suggest to also differentiating between a "task to solve a problem" and a "research task". We are not going to say that the two task types are mutually exclusive. Still there are somewhat different notions that underlie the task types.

The core of a typical task to solve a problem is, a problem that is clearly and unambiguously given. To take an example from experiments with apes, a chimpanzee is hungry (often because of the will of the experimenter), and it has to get the food that is out of its reach. There is here a problem (hunger) and, what is very important, the reward for solving of the problem.

At the basis of a typical research task, there is a need to understand, to grasp, to discover a principle of work, of functioning of something. Thus, a research task by the very nature of it has to include an element of discovering something what earlier was subjectively unknown.

Many experiments that have become classics of behavioral science included a research task into the experimental plan. In the experiments of E. Thorndike (1911) with his puzzle box, a cat was tested if it can understand the functioning of the box to solve the problem (to get out of the box), although it did not come to understanding of the box functioning and showed a sort of trial and error learning. R. Yerkes (1916) forced the apes and monkeys understand according what principle they can get food. The animals have to grasp an abstract idea "first at the left end" to solve the problem (to get the food). The famous DONALD+GERALD problem (Newell, Simon, 1972) contains much of research task in its structure because a subject has to not only discover unknown numerical values for the letters but also discover his/her own way of finding of the values.

Generally speaking, research tasks are mostly novel, ill defined, and, perhaps, poorly rewarded from the environment.

*Semi-binary dialogue*. Information theory postulates that all the living beings exchange signals with their environments. We used a kind of semi-binary dialogue when one side sends a complex signal but the other replies with only "true" or "false". The mentioned above experiments of Yerkes (1916) can be interpreted in terms of semi-binary dialogue because every action of the animals were followed either by getting of the food ("true" signal) or by a sort of discomfort ("false" signal).

*Phase portrait.* A phase portrait is one of the instruments to study a time-dependent process. Suppose that X is the number of errors done by a subject while doing a test task and that there is an obvious time trend of the variable. Thus, for any two subsequent records of X one could calculate a change of X, i.e. its time derivative, with the help standard means. Such a derivative is usually denoted as $\dot{X}$. A phase portrait is then defined as the relationship $\dot{X}=f(X)$ that can help to graphically express the time dynamics of the system being modeled because the value of X will always wane in the range where $\dot{X}<0$ and grow in the range where $\dot{X}>0$. Fig. 1 depicts a schematic example of a phase portrait.

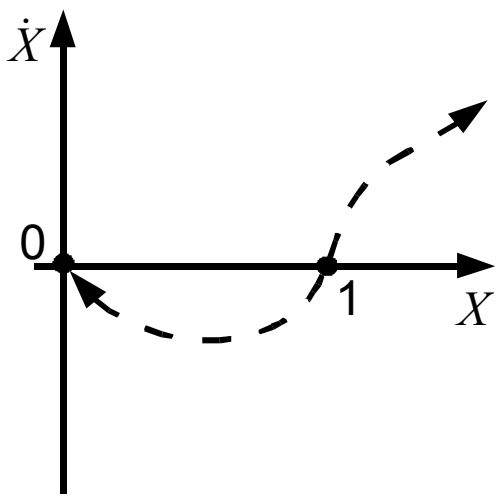

Fig. 1. An example of some hypothetical phase portrait. The relationship

$\dot{X} = f(X)$ is shown as the dashed line. The arrows show the spontaneous dynamics of the system, with the point "0" being stable and the point "1" being unstable steady state.

A description of the research task that was suggested to volunteer participants is given below in the Methods section.

**Methods**

A specially developed computer program Right-Wrong Responder (RWR) was used that generated and visualized the cues material for the participants. The cues material presented geometrical figures as circles, squares, and triangles. Each of the figures had three grades of gray color: light, medium, and dark. Also, they had three grades of size: small, medium, and large. Thus, all the variety of figures consisted of 27 figures variants. The quantity of figures that were shown to the participants as one set equaled nine figures. Every definite set of the cues was generated by the program. A view of a realization of the program work is shown in Fig. 2.

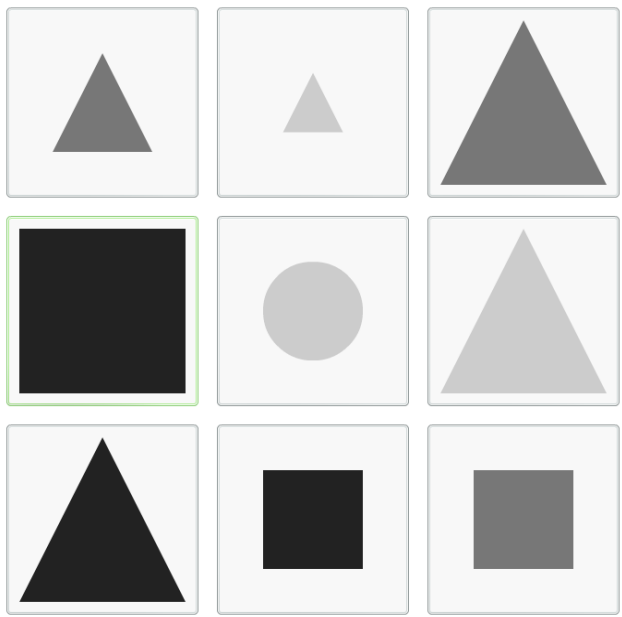

Fig. 2. A typical set of figures generated by the RWR program and shown

to a participant.

The work of a participant with the RWR program was interactive and looked as a semi-binary dialogue. The figures were shown as graphical buttons and the participant was allowed to press any of the figure buttons. As a response to this action, RWR program communicated either "Right choice" or "Wrong choice". The figure "rightness" was defined by an algorithm that was unequivocally programmed in the code. The algorithm was unknown to the participants.

If the participant chose "wrong" figure the RWR program communicated "Wrong choice" and suggested to choose another figure in the same figure set. If the participant chose "right" figure the RWR program communicated "Right choice" and generated another set of figures that the participant was offered to do a new choice in.

The generation of figures was partly random and deterministic: the RWR program guaranteed that there was at least one "right" figure in every set. All other figures were chosen randomly. It means that there was a small probability of generating of figure selections with two or three "right" figures.

The goal that was announced to the participants was to click in such a way that the RWR program always responded "Right choice". It was possible in the only case that the participants understood the algorithm according to which the program assigned the "rightness" of figures: every next "right" figure was different from the previous one. The work with the RWR program finished successfully if the program six times in succession communicated "Right choice", which was considered as the sign of solving of the task. If a participant interrupted the work before the solution it was considered as he/she has not solved the task.

The RWR program was uploaded to a server, and the participants worked completely autonomously through Internet. An English version of the RWR program is available on the page http://sandbox.kspu.ru/en/test1en.html. The data on the clicks by a participant was sent by the program to a file that was individual for every participant. The data were later downloaded and analyzed.

In the following, some statistical properties of the algorithm that generated the cues material are given. If there were only one "right" figure in every set of figures then the mean quantity of "right" figures in a succession of figure sets would equal to unity. Meanwhile, it has been said above that with some probability the program algorithm generated more than one "right" figure.

Our estimations of probabilities for the algorithm to generate various figures in the system of nine figures shows that the probability to generate one "right" figure is 0,735, two - 0,194, tree - 0,056, four - 0,01, five – 0,001. Then, one can calculate that a mean quantity of "right" figures in such a system approximately equals to 1,344, and an estimated mean value of errors done before a random click on a "right" figure equals to 3,38.

The research task was suggested to volunteer participants. About thirty people tried to solve the task, the results of five of which were taken for a detailed analysis. The criteria for selection of participants included all cases where the task was solved under controlled conditions. Also, if the task was not solved there must be evidences that the participant spent enough effort and time to solve the task. The recorded data should not contain lengthy intervals when no activity took place. The conventional identifications and main characteristics of the participants are given in the table.

Table. Basic characteristics of the participants.

| ID of the participants | Time spent for the work, min | Number of clicks done | Success in solving of the task |
|---|---|---|---|
| «K» | 21,1 | 209 | yes |
| «M» | 48,9 | 39 | yes |
| «B» | 13,5 | 83 | yes |

| «Ch» | 16,7 | 71 | no |
| «G» | 14,4 | 219 | no |

## Results and discussion

*Error frequencies*

Figures 3 and 4 show the participants clicks against the time for solvers and non-solvers, correspondingly. These data say of what was the speed of clicking in the course of work with the program. To compare, every figure contains a straight line the slope of which characterizes an average clicking speed for the overall time of work.

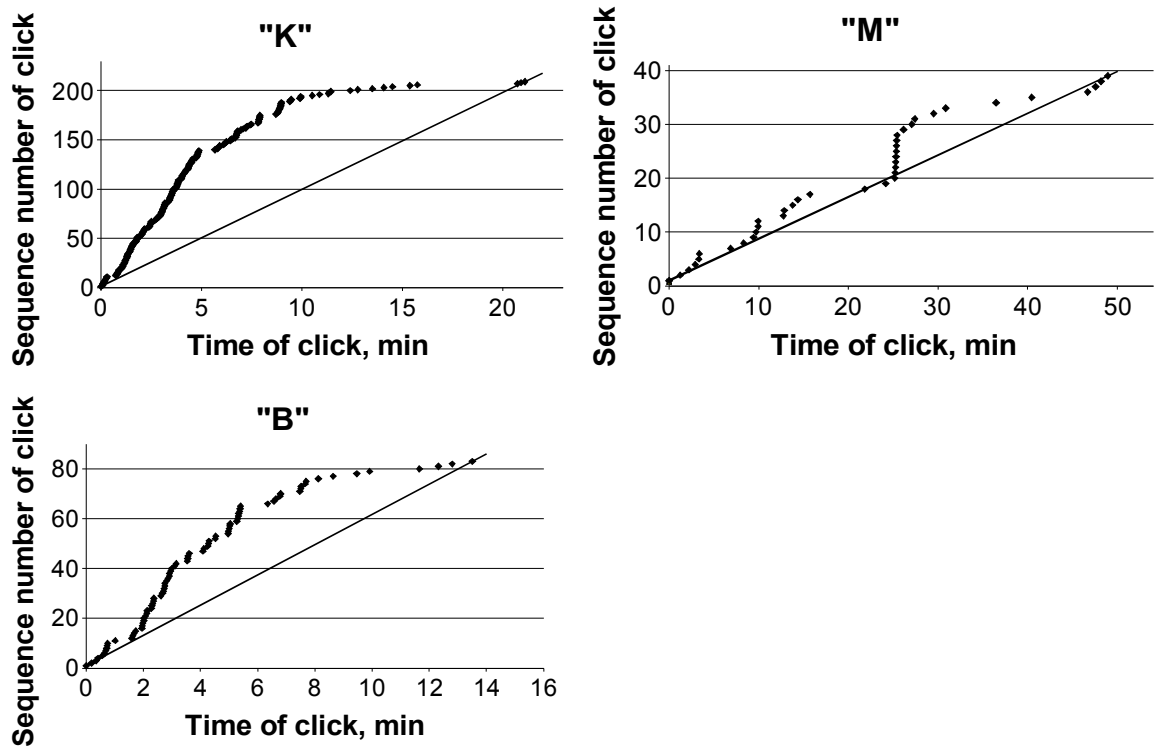

Fig. 3. Sequence number of clicks plotted against time when the clicks were done by solvers. The straight lines characterize the average speed of clicking.

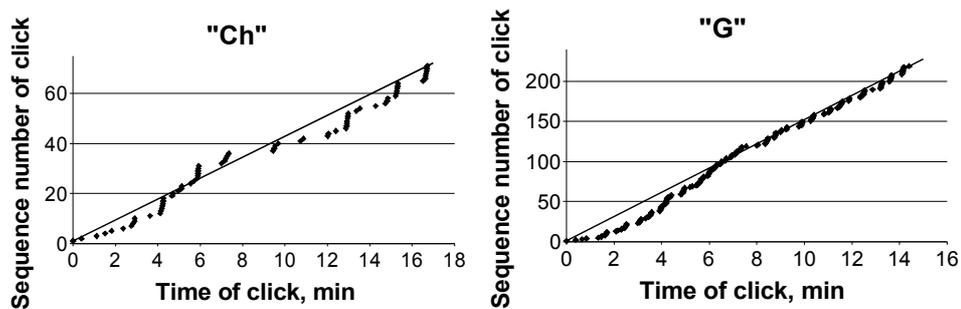

Fig. 4. Sequence number of clicks plotted against time when the clicks were done by non-solvers. The straight lines characterize the average speed of clicking.

A visual analysis of the charts in Figs. 3 and 4 has some meaning for the understanding of the solving dynamics. The data on every participant were divided into a beginning phase and a closing phase. For the solvers, the data were visually divided into a faster and a slower phase, with the portion of the closing slower phase comprising from ⅓ to ¼ from the overall number of clicks. For the non-solvers, because there were no visual references the data were divided into equal beginning and closing phases. These phase data are analyzed below.

The work of the participant with the program was a sort of choosing of figures and receiving of a feedback response from the program. This program feedback may be represented as a sequence of the type of "wwwwRwwwRwwwwwwRwwR..." where "w" denotes "wrong" and "R" denotes "right". If the total number of figures in a given set and a mean number of "right" figures in many sets are known and the choosing of the figures occurs randomly then the lengths of error sequences ("wwww"), i.e. the number of errors between right answers possesses certain statistical properties. In particular, it has been mentioned above that the mean number of errors for this particular algorithm equals to 3,38. This value presents a convenient reference point to compare to numbers of errors made by people participated in the study.

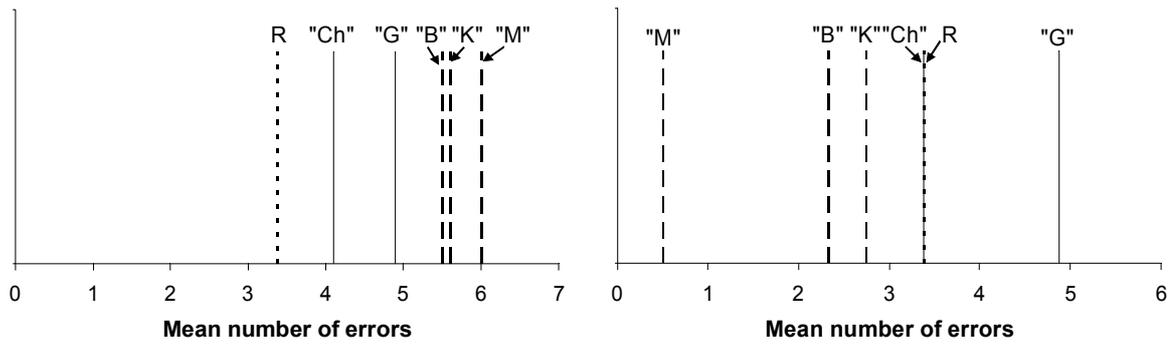

Fig. 5. Mean number of errors made by the participants at the beginning (left) and closing (right) phases of the work. The "R" level denotes the theoretical 3,38 errors for a case of random clicking. The solvers are denoted by dashed lines, and the non-solvers are denoted by solid lines.

Figure 5 gives an idea of the average levels of errors done by the participants at the beginning (left) and closing (right) phases of work with the RWR program. Every vertical segment denotes one participant, and its distance from the reference point corresponds to the calculated mean errors. The theoretical mean error in case of random clicking is denoted by the segment "R".

As it follows from the presented data, the solvers showed bad results of the work at the beginning phase. It means that they did more errors than the non-solvers but also even more that would be in case of random clicking. In particular, the participant "M", who was most effective from the point of view of number of clicks done (see table), showed worst result at the beginning phase.

Unfortunately, small numbers of "right" clicks by participants "M", "B", and "Ch" do not allow us to speak of an estimation of statistical significance of the difference from "R". Using the method of confidence intervals gives us an opportunity to establish the 95%-level difference from "R" only for participants "K" and "G" that did 21 and 20 "right" clicks, correspondingly, at the beginning phase.

The closing phase looks radically different from the beginning phase (Fig.

5, right). All the solvers show visibly better results, i.e. less errors than it would be at random clicking. It is important to note that the estimations of mean errors for solvers were done without taking into account the final series of error-free clicks. The conditions of the task (see Methods section) were so that the final series of six "right" clicks were supposed be in a succession, without any error clicks between them. Therefore, the data of solvers have always a final series of zeros. Taking the zeros into the calculation would make the estimated mean errors even lower. Thus, the mean errors of solvers correspond to the time of work that immediately preceded the successful solution.

The non-solvers show the results at the random level or worse than that at the closing phase (Fig. 5, right, "Ch" and "G").

Solving of every task requires definite resources. The results presented here allow one to suppose that solving of research tasks in highly uncertain environment requires some psychological resources. At least, those who endeavor to solve research tasks have to have a sort of "right to mistake", which gives one opportunity to try out and evaluate many solution variants most of which are erroneous or lead to a dead end. Sometimes it is more important if the challenger finally solves the task, not the amount of mistakes he/she does, especially at the beginning.

*Phase portrait of solution*

The phase portraits are powerful instrument of preliminary analysis and interpretation of data. In particular, it was shown that a small amount of reasonable assumptions allowed one to construct a typology of learning individuals (Gavrikov, Khlebopros, 2009). We might expect that one of the suggested in the cited study types, self-learning, would describe the course of the task solving for the solvers considered above. A theoretical phase portrait of a solver under the given condition of the research task may look as in Fig. 6. The portrait in Fig. 6 says that a person begins with some amount of errors N and then smoothly makes his/her work better and better doing less and less errors. It brings him/her persistently closer and closer to the stable zero point. The zero point means in our context that the person stops to do errors.

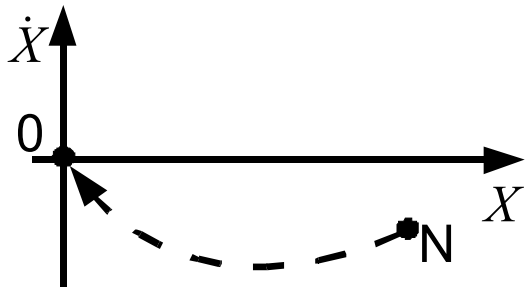

Fig. 6. A theoretical phase portrait of self-learning type (Gavrikov, Khlebopros, 2009) for the case when $X$ stands for number of errors. $\dot{X}$ denotes the change in error number with time, N stands for an initial number of errors. The point "0" is a steady stable state.

The analysis of the data received shows a more complicated picture that a theoretical phase portrait (Fig. 6). The real phase portraits for solvers are shown in Fig. 7. In the figure, the flow of time is marked by small arrows. It is obvious that no monotonous flow from many errors to zero errors is observed. The dynamics of task solving resembles rather spiral oscillations with a definite trend to lower errors. The spiral trajectory for the participants "K" and "B" takes place both in the beginning and closing phase of solving. The phase portrait for the participant "M" has only one cycle.

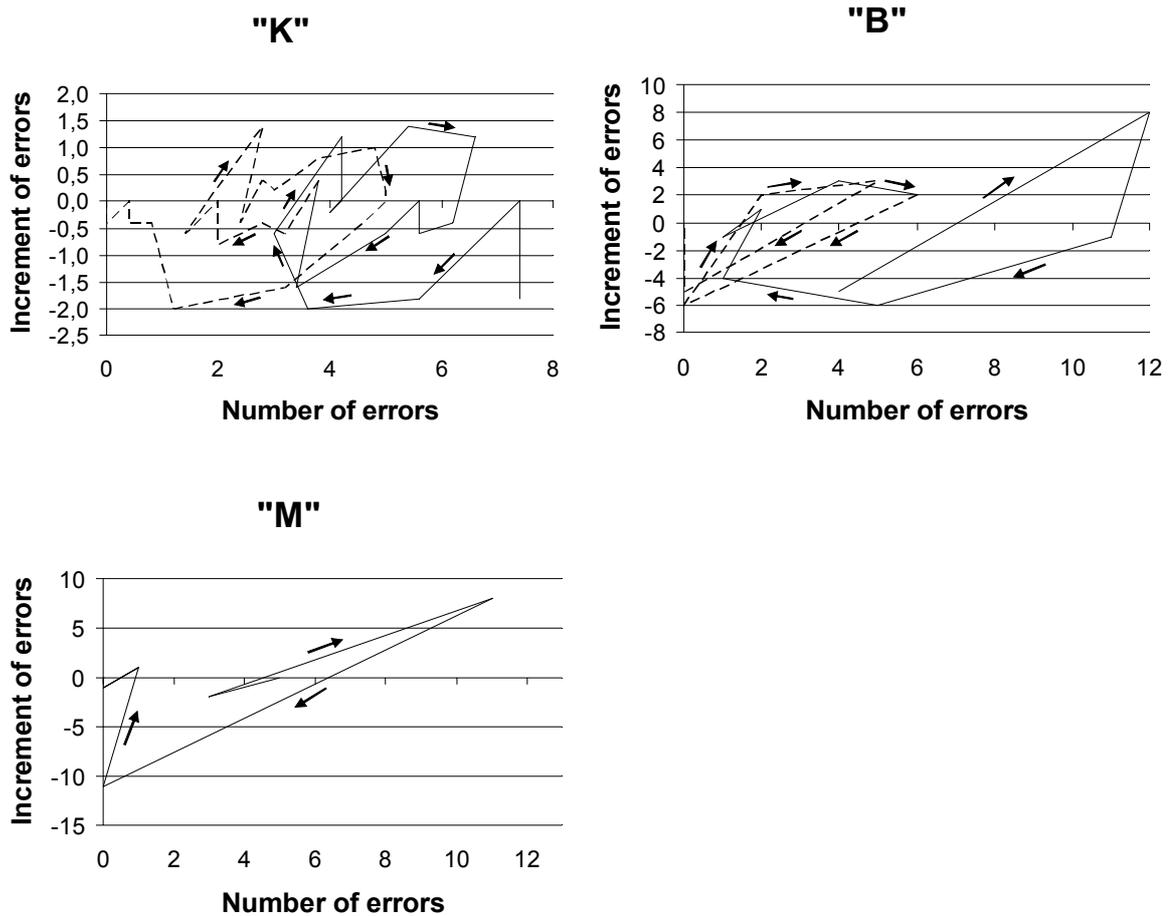

Fig. 7. Phase portraits of the work with the RWR program for solvers. Because of the large number of clicks, the data for the participant "K" were smoothed by a sliding mean with the window width of 5. The time flow is marked by arrows. For the participants "K" and "B", the beginning phase is shown as the solid line and the closing phase is shown as the dashed line.

The phase portraits for non-solvers are shown in Fig. 8. As it is seen from the data, the trajectories of non-solvers are also of the sort of spiral cycles. However while working with the RWR program the participants "Ch" and "G" either did not move to the solution or moved too slow. The estimations of mean increments of error numbers give an idea of difference between solvers and non-solvers. The mean increments for the participants "K", "M", and "B" amount -0.27, -0.56, and -0.69 correspondingly, i.e. they are negative, which means that on average the number of errors dropped. In the same time, the mean increments for the participants "Ch" and "G" were 0.21 and 0.17, i.e. positive.

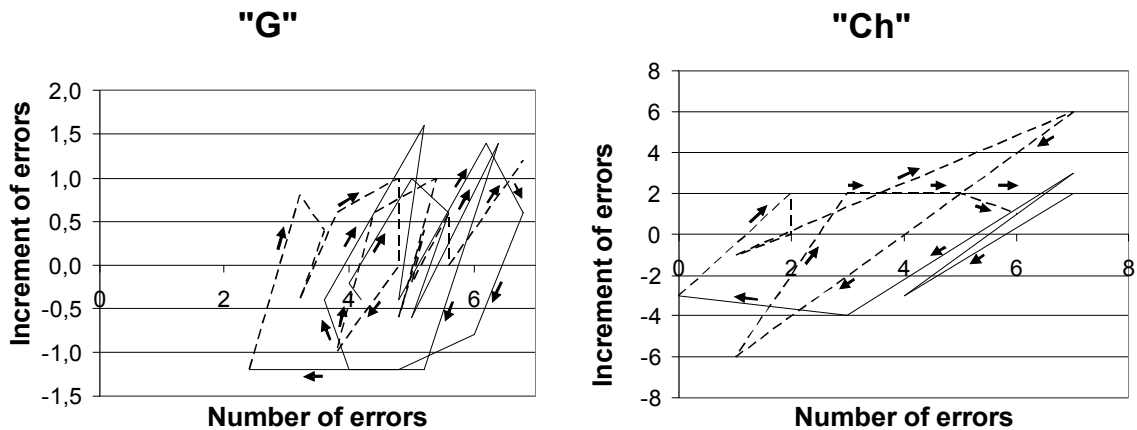

Fig. 8. Phase portraits of the work with the RWR program for non-solvers. Because of the large number of clicks, the data for the participant "G" were smoothed by a sliding mean with the window width of 5. The time flow is marked by arrows. The beginning phase is shown as the solid line and the closing phase is shown as the dashed line.

Therefore, the spiral cyclic course of solving process, which presumably reflects the mental work of a human, is a common characteristic for both solvers and non-solvers. The data presented allow us to propose a possible source of the spiral cyclic phase portraits of the participants. In other words, we would like to propose a mechanism in terms of thinking process that underlies the observed spiral cyclic dynamics.

Fig. 9 gives a generalized picture that describes a successful solving process of the research task. Supposedly, those parts of the cycles where $\dot{X}<0$ reflect the mental process of putting forward a hypothesis by the human. The hypothesis may be both conscious and subconscious and is about how the human should act. All the hypotheses that generate the solvers except the last one are only partially valid. The validity of a hypothesis allows the human to reduce the number of errors within a definite time interval. However a partially valid hypothesis obviously has limitations and if the person insists to follow the hypothesis he/she would do more and more errors. The parts of the cycles where $\dot{X} > 0$, i.e. the number of errors grows, are shown as dashed lines.

Rethinking of failures may lead the person to that a new, closer to the truth, hypothesis is maturing, and the cyclic dynamics comes to the success when the person generates a hypothesis that happens to be completely valid and leads him/her to the solution.

The notion that the thinking process in humans goes by a sort of oscillations appeared as early as late 19th century in the work "The principles of psychology" (1890) by one of the founders of the modern psychology William James. Observations of oscillation phenomena in many natural sciences suggest that it is an effect of time lagging that underlie every cyclic movement. In our case, the time lagging may be so that the person fails to generate a more valid hypothesis at the right time when the old one reaches its applicability limits. Langley and Rogers (2005) made a critical review of the classical theory of problem solving. In particular, it was noted that eager execution of partial plans can lead the problem solver into physical dead ends (Langley, Roger, 2005).

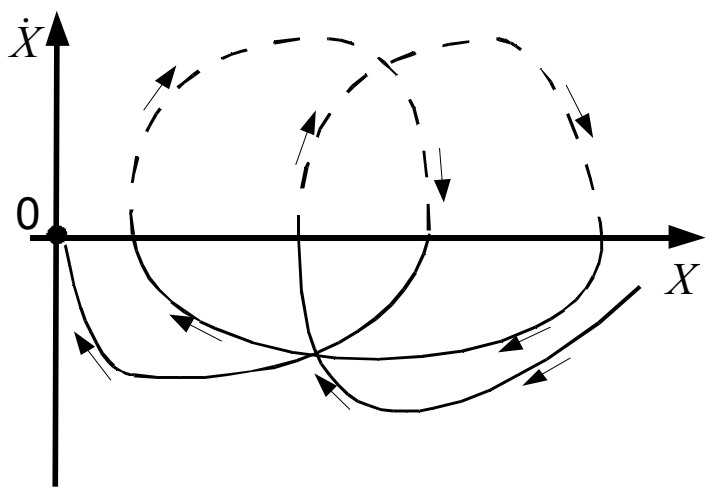

Fig. 9. A generalized phase portrait of a successful solution of the research

task. *X* stands for the number of errors. $\dot{X}$ stands for changes of the number of errors. The solid lines denote those segments of the trajectory where the number of errors reduce, supposedly, due to a partially valid hypothesis. The dashed lines denote those parts of trajectory where the number of errors grows because, supposedly, the partially valid hypothesis is no longer valid.

To conclude the description of results, we should mention two important points, one of them relates to the interpretations of the results and the other touches the methodology of the study as a whole. First, it may look like the research task technique used in the study gives an estimation of intellectual capacities of a participant. Even if it may be the case it should be taken with caution. Making grounded conclusions on the matter of intellect requires, at least, more research because intellect is a rather controversial issue in psychology. Rather, the technique provides a complex picture of the capacity to solve the particular research task. This picture consists of many components, which does not allow us to associate the failure to solve the task exclusively with the intellectual component. In particular, there is an impression that motivation is one of the most important sources of the capacity to solve the research task. To successfully solve the task one should have a motivation, at least in the form of simple curiosity. Some participant that cannot be suspected as having low intellect did not solve the task just because they did not realize why they have to be occupied with all the stuff.

Second, as it follows from the technique description we considered the number of errors as the main parameter of learning progress. A regular absence of errors in solving of a class of tasks means, in our viewpoint, that the participant understands, or realizes, the structure of the task and the algorithm that can lead to success.

This approach is not indisputable. For example, Shuikin and Levshina (2008) give the following argument. If a computer program solves a class of mathematical tasks error-free then we might conclude that the program "understands" mathematics, which is false. Therefore, in accordance with the author's idea, reduction in the error number or absence of them cannot be a sign of task understanding.

In our view, this argument is a rather lame one because the error-free solving of mathematical tasks by such a program witnesses that the authors of the program understand the tasks. A computer program like MathLab is not an independent living being that is capable of learning. However, the program authors do possess the capacities to learn mathematics.

We believe that within the definite case considered here the dynamics of error numbers can be taken as a measure of learning of a living being that do not have initial specific knowledge of the task and its solution. The highest level of learning is achieved in the case of a deep understanding of a task structure that gives one a capacity not to do errors at all.

**Conclusion**

Studies of thinking dynamics during solving of tasks are of considerable interest. The reason for the interest is that the non-solvers may belong to at least two categories. The first one are those people who cannot solve a particular task just because its complexity goes beyond their abilities. The second category are those who do not solve tasks because of secondary causes produced by, for example, artificial learning environments. It may happen that there is a lack of time while the person is of a slow reflective type. Or, an impulsive person has to solve tasks under interfering influence of other highly priority events. At last, some people may quickly spend their nervous resources and require some extra attempts. In fact, there may be various circumstances that prevent a person from successful solving, "now and here".

We believe that it would be important to have methods to differentiate between the two categories of individuals. To be in a position to do it one should have a more profound understanding of the solving dynamics so that the success in solving may be foreseen before it is achieved. Hypothetically, a particular kind of behavior, a research behavior, can be defined and studied. If a person shows a research behavior in respect to a class of tasks we could predict the success long before it occurs.

**Acknowledgments**

The server space for the RWR program, as well as the programming software were provided by Krasnoyarsk State Pedagogical University. We are grateful to all the volunteers who agreed to test our method and to Sergej Glasner for the consultations in programming.